%% file: 01_main.tex
\documentclass[10pt,twocolumn,letterpaper]{article}
\usepackage{titling}

\input{macro.tex}

\usepackage{cvpr}              %

\usepackage{graphicx}
\usepackage{amsmath}
\usepackage{amssymb}
\usepackage{booktabs}
\usepackage{url}
\usepackage{amsfonts}

\usepackage[pagebackref,breaklinks,colorlinks]{hyperref}

\usepackage[capitalize]{cleveref}
\crefname{section}{Sec.}{Secs.}
\Crefname{section}{Section}{Sections}
\Crefname{table}{Table}{Tables}
\crefname{table}{Tab.}{Tabs.}

\def\cvprPaperID{8292} %
\def\confName{CVPR}
\def\confYear{2022}

\begin{document}

\title{TransforMatcher: Match-to-Match Attention for Semantic Correspondence}

\author{Seungwook Kim \hspace{0.8cm} Juhong Min \hspace{0.8cm} Minsu Cho \vspace{1.5mm} \\
Pohang University of Science and Technology (POSTECH), South Korea \vspace{1.5mm}\\
\small
\href{http://cvlab.postech.ac.kr/research/DHVR}{\url{http://cvlab.postech.ac.kr/research/TransforMatcher}}
}
\maketitle

\begin{abstract}

Establishing correspondences between images remains a challenging task, especially under large appearance changes due to different viewpoints or intra-class variations.
In this work, we introduce a strong semantic image matching learner, dubbed \textit{TransforMatcher}, which builds on the success of transformer networks in vision domains. 
Unlike existing convolution- or attention-based schemes for correspondence, TransforMatcher performs global match-to-match attention for precise match localization and dynamic refinement. 
To handle a large number of matches in a dense correlation map, we develop a light-weight attention architecture to consider the global match-to-match interactions. 
We also propose to utilize a multi-channel correlation map for refinement, treating the multi-level scores as features instead of a single score to fully exploit the richer layer-wise semantics.
In experiments, TransforMatcher sets a new state of the art on SPair-71k while performing on par with existing SOTA methods on the PF-PASCAL dataset.

\end{abstract}

\input{sections/1_introduction}
\input{sections/2_relatedwork}

\input{sections/3_Preliminary}

\input{sections/4_method}

\input{sections/5_experiments}
\input{sections/6_conclusion}

\vspace{2mm}
\noindent
\textbf{Acknowledgement.} 
This work was supported by Samsung Advanced Institute of Technology (SAIT) and also by the NRF grant (NRF-2021R1A2C3012728) and the IITP grants (No.2021-0-02068: AI Innovation Hub, No.2019-0-01906: Artificial Intelligence Graduate School Program at POSTECH) funded by the Korea government (MSIT). 

{\small

\input{01_main.bbl}
}
\input{02_supp}

\end{document}

%% file: macro.tex
\usepackage{color}
\usepackage{mathtools}
\usepackage{amsmath}
\usepackage{bbm}
\usepackage{multirow}
\usepackage{booktabs}
\usepackage{amsfonts,amssymb}
\usepackage{pifont}
\usepackage{xcolor,colortbl}
\usepackage{wrapfig}
\usepackage{mwe}

\definecolor{Gray}{gray}{0.85}

\newcommand{\Eq}[1]  {Eq.\ (\ref{eq:#1})}

\newcommand{\Fig}[1] {Figure \ref{fig:#1}}

\newcommand{\Tbl}[1]  {Table \ref{tbl:#1}}

\definecolor{purp}{rgb}{0.65, 0.16, 0.65}
\definecolor{green}{rgb}{0, 0.5, 0}

\newcommand{\blue}[1]{{\color{blue}{#1}}}

%% file: sections/1_introduction.tex
\section{Introduction}

Establishing correspondences between images is a fundamental task in computer vision, and is used for a wide range of problems including 3D reconstruction, visual localization and object recognition~\cite{forsyth:hal-01063327}.
With the recent advances of deep neural networks, many learning-based keypoint extractors and feature descriptors were introduced~\cite{detone2018superpoint, tian2019sosnet, truong2020glunet, revaud2019r2d2, dusmanu2019d2}, showing significantly improved performances over their traditional counterparts~\cite{lowe1999sift, lowe2004sift, dalal2005histograms, bay2006surf}.
More recently, dense feature matching methods - which use all extracted features for matching - have shown impressive performances despite higher computation complexities~\cite{rocco2018neighbourhood, li20dualrc, min2021chm}.
However, establishing reliable correspondences between images under the presence of intra-class variations \ie, different instances of the same category, remains a critical challenge for semantic visual correspondence ~\cite{min2020dhpf, liu2020semantic, min2021chm, rocco18weak, rocco2018neighbourhood, huang2019dynamic, min2019hyperpixel, kim2018recurrent, jeon2020guided, cho2021semantic, han2017scnet, ham2016proposal, ham2018proposal, jeon2018parn, truong2020glunet, rocco17geocnn, liu2020semantic}. 

\input{figures/teaser}

The idea of applying high-dimensional convolutional layers on the 4D feature correlation map was first proposed in NCNet~\cite{rocco2018neighbourhood}, which proposes that unique matches will support the nearby ambiguous matches.
Among the various methods proposed for establishing semantic correspondences, NCNet and its follow-up methods have shown impressive results~\cite{rocco2018neighbourhood,rocco2020sparse,min2021chm,huang2019dynamic,li2020correspondence}.
These methods evidence that considering the match-to-match consensus by utilizing the full set of dense correspondences represented by the 4D correlation map is effective in establishing robust and accurate semantic correspondences.
However, the convolution-based methods suffer from inherent limitations of {\em local} and {\em static} transformations; performing the same local transformation over all spatial positions of the input.

While convolutional neural networks have been the de-facto standard for visual correspondence, transformer networks have recently shown promising results in the computer vision domain.
The success of transformer networks can be largely attributed to their {\em dynamic} feature transform unlike stationary convolutional layers, and the {\em non-local} interactions between input elements which enable easy scalability to attend to global contexts.
For example,ViT~\cite{dosovitskiy2020vit} attains excellent results compared to convolutional baselines on the task of image recognition with fewer training computational resources; Segmenter~\cite{strudel2021} outperforms convolution-based methods by modeling global context already at the first layer and throughout the network. 
These pioneering work show that transformer layers are attractive alternatives to convolutional layers in vision models.

Inspired by the effectiveness of match-to-match consensus consideration and transformer networks, we propose a novel semantic matching pipeline, dubbed {\em TransforMatcher}.
Specifically, we introduce match-to-match attention, a self-attention based mechanism to consider the \emph{global} match-to-match interactions by leveraging the 4D correlation maps computed from features of images to match.
Considering the global match-wise interactions allows to capture long-range relevance across matches, and incorporates geometric consistency between distant matches in a dynamic manner especially under challenging appearance variations.
This is achieved by considering each spatial entry of the 4D correlation map (\ie a match) as an individual element for attention, which differs from LoFTR~\cite{sun2021loftr} or CoTR~\cite{jiang2021cotr} which consider the patch-to-patch relations within or across 2D feature maps through self- or cross-attention.
\Fig{teaser} visualizes the comparison between patch-to-patch and match-to-match attention.

Our contributions can be summarized as follows:
\begin{itemize}
    \item We propose the TransforMatcher, a novel image matching pipeline built on transformer networks for dynamic match-to-match interactions at a global scale, 
    \item To the best of our knowledge, we are the first to model the {\emph global} interactions between the full set of dense correspondences using a self-attention mechanism within feasible computational constraints,
    \item We leverage multi-level correlation scores to be used as features, improving over using a single score,
    \item We demonstrate state-of-the-art or on-par performances on standard benchmarks of category-level matching - SPair-71k and PF-PASCAL.
\end{itemize}

\input{figures/method_comparison}

%% file: figures/teaser.tex
\begin{figure}
    \centering    
    \begin{center}
        \includegraphics[width=0.99\linewidth]{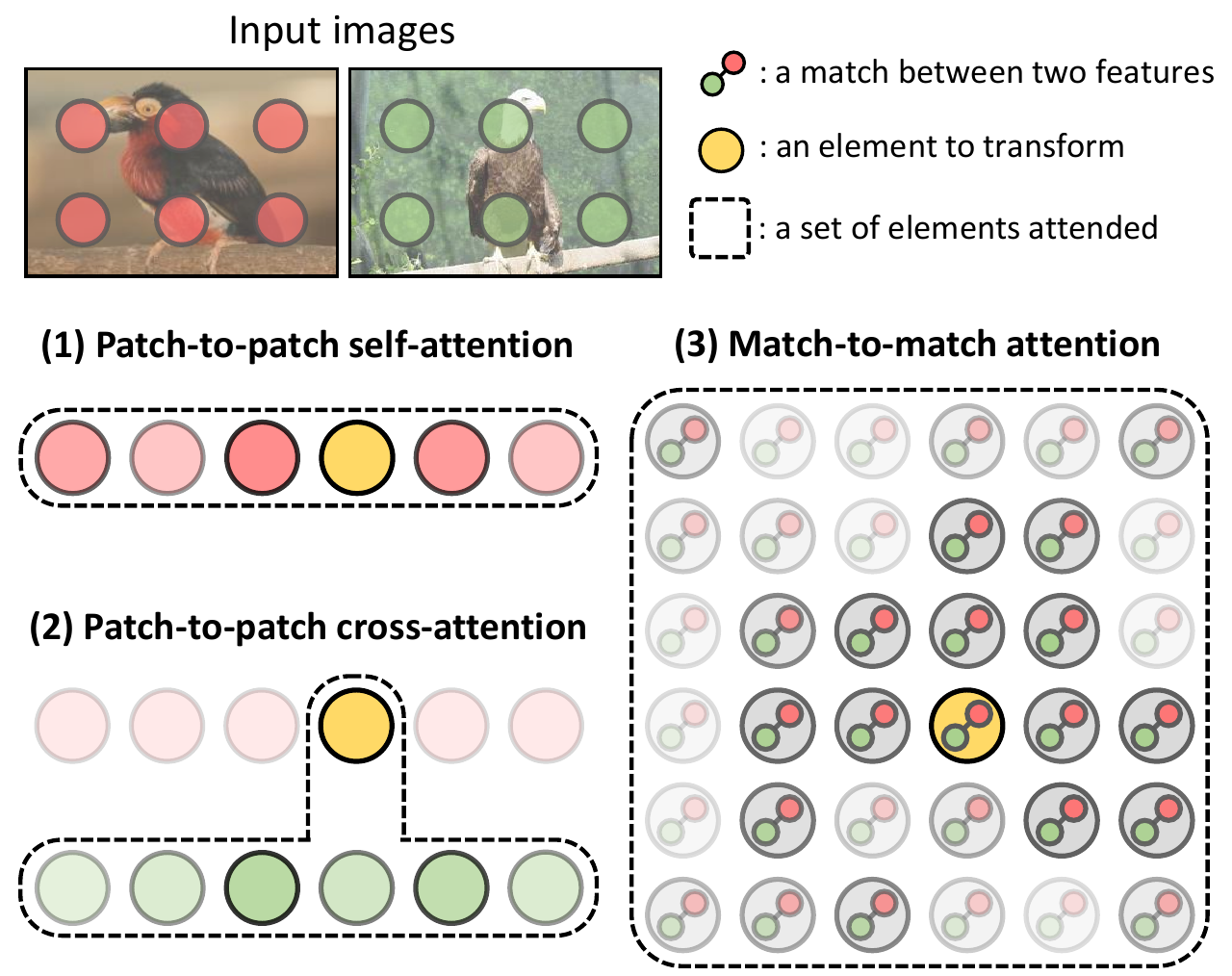}
    \end{center}
    \vspace{-5.0mm}
      \caption{\textbf{Patch-to-patch vs. Match-to-match attention.} Patch-to-patch attention considers each position in a 2D feature map as an individual element, while match-to-match attention considers every match in pair-wise correlations as an individual element.
}
    \vspace{-5.0mm}
\label{fig:teaser}
\end{figure}

%% file: figures/method_comparison.tex
\begin{figure}
    \centering    
    \begin{center}
        \includegraphics[width=0.99\linewidth]{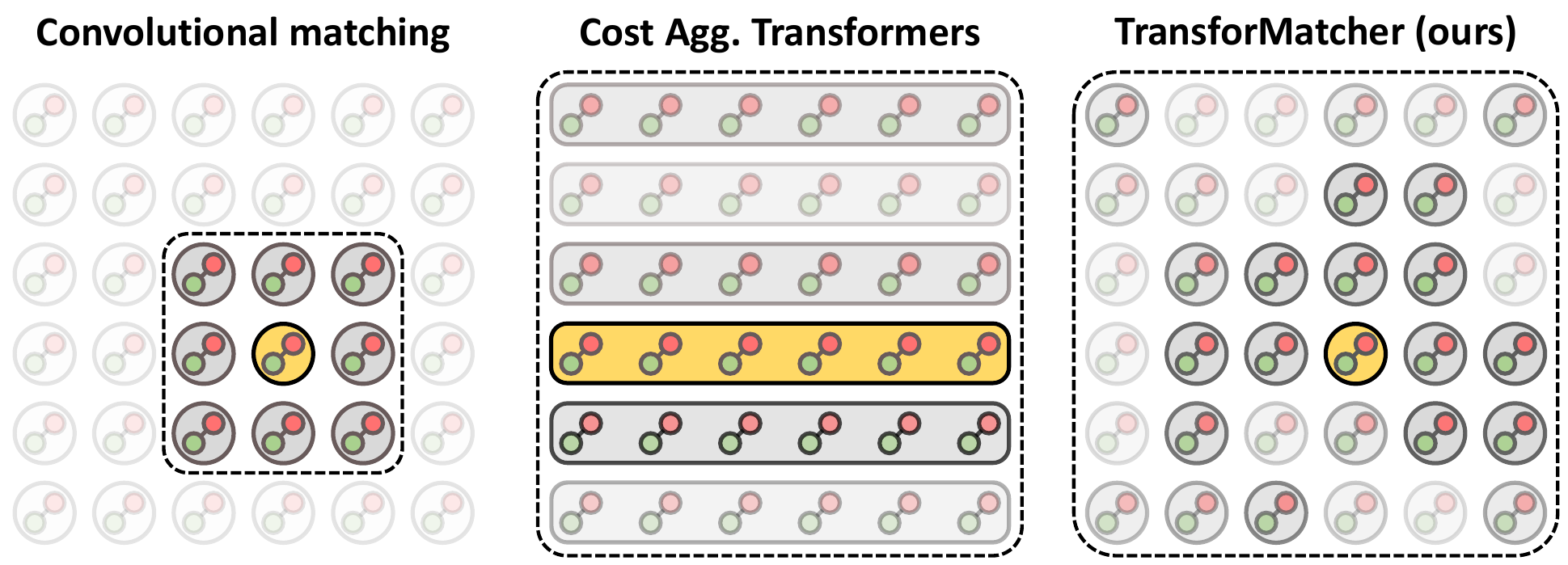}
    \end{center}
    \vspace{-5.0mm}
      \caption{\textbf{Conceptual difference between recent methods and ours.} Convolution-based matching methods~\cite{rocco2018neighbourhood, min2021chm, huang2019dynamic, min2021cpchm} (left), Cost Aggregation Transformers~\cite{cho2021semantic} (middle), and ours (right).}

    \vspace{-5.0mm}
\label{fig:method_comparison}
\end{figure}

%% file: sections/2_relatedwork.tex
\input{figures/overview_architecture}

\section{Related work}

\smallbreak
\noindent \textbf{Category-level matching using convolutional networks.}
Category-level matching, a.k.a. semantic matching aims to find corresponding elements between images of different instances in the same category.
Traditional approaches to category-level matching use hand-crafted descriptors to obtain matches between images~\cite{cho2015unsupervised,taniai2016joint}.
Recent approaches~\cite{min2020dhpf, li2020correspondence, jeon2020guided} build on the success of deep learning to extract learned features from convolutional neural networks, usually pretrained on the ImageNet classification task~\cite{krizhevsky2012imagenet}.
An emerging trend is to exploit high-dimensional convolution on the correlation map obtained from features of images to match, considering the local match-to-match consensus to refine the correlation map~\cite{rocco2018neighbourhood, lee2020pmnc, lee2021distill, min2021chm}.

While these work have proven the efficacy of utilizing correlation maps for \textit{local} match-to-match consensus in discovering reliable matches, we propose that exploiting the \textit{global} match-to-match interactions further enables to capture long-range relevance between matches, which is crucial for image pairs with challenging appearance variations.
We therefore impose efficient match-to-match attention on the 4D correlation map, exploiting a lightweight attention scheme to easily scale to use the global context.

\smallbreak
\noindent \textbf{Image matching using transformer networks.}
Following the success of transformer networks in computer vision~\cite{dosovitskiy2020vit, pmlr-v139-touvron21a, Vaswani_2021_CVPR, liu2021swin, wang2021pyramid}, recent instance-level matching methods propose to use transformer networks.
On a conceptual level, SuperGlue~\cite{sarlin20superglue} employs an attention-like mechanism on a set of sparse keypoints and their descriptors.
LoFTR~\cite{sun2021loftr} extends this idea to dense 2D feature maps of the images to match, leveraging self- and cross-attention layers between the feature maps to generate strong features for matching.
COTR~\cite{jiang2021cotr} concatenates the feature maps of images to match along the spatial dimension, which is used as input to the transformer networks together with the query point to output the target point.
Note that these methods are actually performing patch-to-patch attention, not leveraging the match-to-match interactions between feature maps.

The work of CATs~\cite{cho2021semantic} does employ the transformer networks to model global consensus on the 4D correlation map for the task of semantic correspondence.
However, they differ from our work in the following aspects: 
(1) We use every match on the correlation map as the input element and multi-level scores as features to perform match-to-match attention to model fine-grained interaction, but CATs reshapes the 4D correlation map to 2D feature maps to perform patch-to-patch attention, modeling a comparatively coarse-grained interaction between elements. This is illustrated in \Fig{method_comparison}.
(2) CATs additionally concatenates a transformed feature map to the reshaped correlation map, increasing the memory overhead of each transformer layer, making it infeasible to stack multiple layers.

\smallbreak
\noindent \textbf{Efficient Transformers.}
Due to the quadratic complexity of conventional transformers~\cite{vaswani2017attention}, they are infeasible to model extremely long-range interactions.
This motivates the use of efficient transformers with lower computational complexity for feasible computation overhead when handling long sequences.
Reformer\cite{kitaev2020reformer} reduces the complexity down to log-linear using locality-sensitive hashing and reversible residual layers.
Linformer\cite{wang2020linformer} approximates the self-attention mechanism using low-rank matrices for linear complexity.
Instead of relying on sparsity or low-rankedness, Performer\cite{choromanski2020rethinking} proposes positive orthogonal random features approach (FAVOR+) to achieve linear complexity as well.
Recently, Fastformer\cite{wu2021fastformer} proposes an architecture which uses additive attention techniques only with element-wise products.
We build on the success of additive attention to implement global match-to-match attention for its scalable complexity and efficacy.

%% file: figures/overview_architecture.tex
\begin{figure*}
    \begin{center}
        \includegraphics[width=1.0\linewidth]{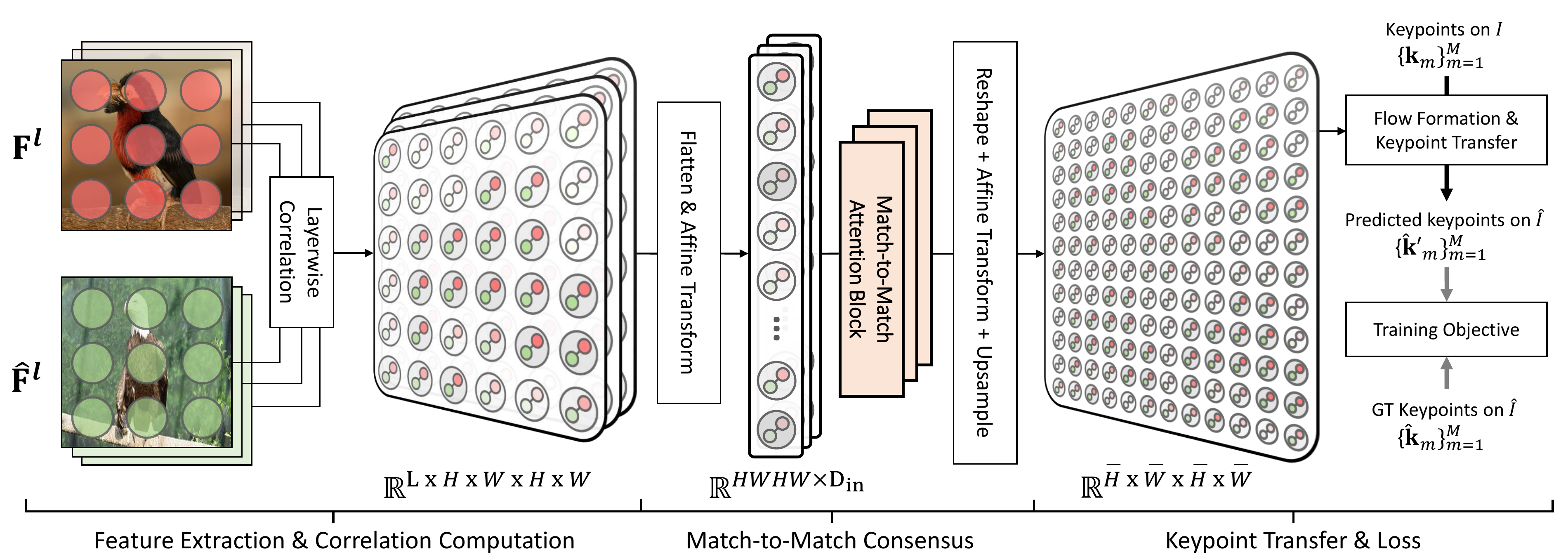}
    \end{center}
    \vspace{-3.0mm}
      \caption{\textbf{Overview of TransforMatcher.}
      The feature maps extracted from an image pair are used to compute a multi-channel correlation map to be processed by our match-to-match attention module for refinement.
      We construct a dense flow field from the resulting correlation map, which can be used to transfer keypoints for training with keypoint pair annotation.
    }
    \label{overview_architecture}
    \vspace{-4.0mm}
\end{figure*}

%% file: sections/3_Preliminary.tex
\input{figures/fastformer_detail}
\section{Preliminaries: Transformer}

Transformers~\cite{vaswani2017attention} are built on multi-head self-attention (MHSA) which consists of multiple self-attention layers.
Each self-attention layer takes input elements $\mathbf{X} \in \mathbb{R}^{T \times D_{\text{in}}}$ to form global self-attention matrices using linear projections of $\mathbf{W}^{(h)}_\mathrm{Q}, \mathbf{W}^{(h)}_\mathrm{K} \in \mathbb{R}^{D_{\text{in}} \times D_{h}}$ and $\mathbf{W}^{(h)}_\mathrm{V} \in \mathbb{R}^{D_{\text{in}} \times D_{v}}$, capturing long-range dependencies between the elements: 
\begin{align}
\label{eq:vanilla_attention}
    \text{SA}^{(h)}(\mathbf{X}) &= \sigma(\tau\mathbf{XW}^{(h)}_\mathrm{Q}( \mathbf{XW}^{(h)}_\mathrm{K})^\top)\mathbf{XW}^{(h)}_\mathrm{V} \\
    &= \sigma(\tau\mathbf{Q}^{(h)} \mathbf{K}^{(h)\top})\mathbf{V}^{(h)},
\end{align}
where $(h)$ is the head index, $\tau$ is a scaling parameter, and $\sigma(\cdot)$ is row-wise softmax function. 
The MHSA layer with $N_h$ heads aggregates the self-attention outputs by affine transformation with $\mathbf{W}_{O} \in \mathbb{R}^{N_h D_v \times D_{\text{out}}}$ and $\mathbf{b}_{O} \in \mathbb{R}^{D_{\text{out}}}$:
\begin{align}
    \label{eq:MHSA}
    \text{MHSA}(\mathbf{X}) = \underset{h \in [N_h]}{\text{concat}}\big[\text{SA}^{(h)}(\mathbf{X})\big]\textbf{W}_{\mathrm{O}} + \mathbf{b}_{\mathrm{O}}.
\end{align}
It can be seen that the computational complexity of the transformer architecture is quadratic with respect to the sequence length $T$, being a fundamental bottleneck when handling long sequences ($T \gg D_h$).
This bottleneck also pertains to our case of processing 4D correlation map, \ie, a full set of pair-wise correlations between two 2D feature maps, as establishing match-to-match attention matrix in self-attention layer demands {\em quartic} memory with respect to the spatial size of the feature maps.
In the next section, we provide an overview of our method as well as an efficient self-attention layer which implements global match-to-match interactions without quartic complexity.

%% file: figures/fastformer_detail.tex
\begin{figure*}
    \begin{center}
        \includegraphics[width=1.0\linewidth]{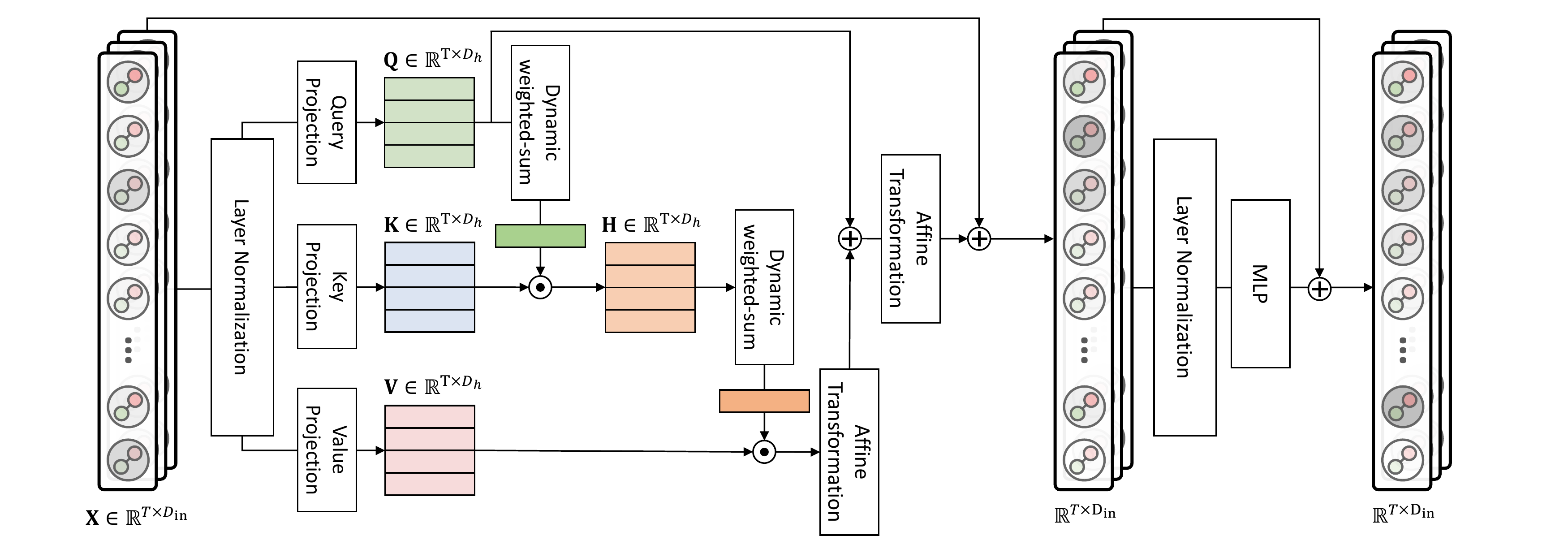}
    \end{center}
    \vspace{-4.0mm}
      \caption{\textbf{Match-to-match attention module.} 
      The multi-channel correlation map is projected to query, key and value matrices, which are multiplied with rotary positional embeddings.
      The match-to-match attention module exploits additive addition mechanisms to aggregate query/key matrices to global vectors, which is used for element-wise product to induce global context awareness. 
      The final output is projected to a single-width channel to be reshaped to a refined 4D correlation map.
}
\label{fig:fastformer_detail}
    \vspace{-4.0mm}
\end{figure*}

%% file: sections/4_method.tex
\section{TransforMatcher}

We first provide an overview of our TransforMatcher pipeline.
Given a pair of images to match, a feature extractor provides a set of intermediate feature map pairs which are used to construct a multi-channel correlation map.
Due to multifarious match-wise interactions within the 4D global correlation map, we employ additive attention with linear complexity to perform match-to-match attention with feasible computation overhead.
We refine the multi-channel correlation map with several match-to-match attention layers, considering the global context within the correlation map in a dynamic manner.
The refined correlation map is used to construct a dense flow field, which can be used for keypoint transfer to supervise our pipeline with ground-truth keypoint pair annotations.
Fig.~\ref{overview_architecture} illustrates the overview architecture of our method.

\subsection{Multi-channel correlation computation}
We use the ImageNet-pretrained ResNet-101~\cite{he2016deep} architecture as the feature extractor.
We use all bottleneck layers of \texttt{conv4\_x} and \texttt{conv5\_x} to extract the features given an input pair of images $I, \hat{I} \in \mathbb{R}^{H \times W \times 3}$, and denote the set of intermediate feature pairs as $\{(\mathbf{F}^{l}, \hat{\mathbf{F}}^{l})\}_{l=1}^{L}$.

A feature map pair extracted from the same bottleneck layer, $\mathbf{F}^l, \hat{\mathbf{F}}^l \in \mathbb{R}^{H_l \times W_l \times D_l}$, are used to construct a correlation map $\mathbf{C}^l \in \mathbbm{R}^{H_l\times W_l \times H_l \times W_l}$ which represents the confidence score for all candidate correspondences between the two feature maps.
Given a set of feature map pairs from different bottleneck layers $\{(\mathbf{F}^l, \hat{\mathbf{F}}^l)\}_{l=1}^L$, we compute the 4D correlation tensors for each pair as follows:
\begin{align}
\mathbf{C}^l_{\mathbf{x}, \hat{\mathbf{x}}} = \text{ReLU}\Big( \frac{\mathbf{F}^l_{\mathbf{x},:} \cdot \hat{\textbf{F}}^l_{\hat{\mathbf{x}},:}}{\|\mathbf{F}^l_{\mathbf{x},:}\|\| \hat{\mathbf{F}}^l_{\hat{\mathbf{x}},:} \|} \Big),
\end{align}
where $\mathbf{x}$,$\hat{\mathbf{x}} \in \mathbb{R}^{2}$ refer to 2-dimensional spatial positions of the feature maps corresponding to the image pair $(I,\hat{I})$.
The $L$ correlation tensors are then stacked together along the channel dimension after bilinear interpolation to the size of $H \times W \times H \times W$, \ie, $\frac{1}{16}$ the size of the input image resolutions, resulting in the final multi-channel correlation map $\mathbf{C} \in \mathbbm{R}^{L \times H \times W \times H \times W}$.

This is unlike correlation maps used in prior work~\cite{rocco2018neighbourhood}, which only have a single channel, \ie, one similarity score value for each pair of positions between the source and target feature maps.
By constructing a multi-channel correlation map, we treat the multi-level scores for each candidate match as \textit{features} instead of a single \textit{score}.
This leverage of different correlation tensors across the bottleneck layers allows us to exploit the richer semantics in different levels of feature maps, unlike previous methods which disregard the layer-wise similarities and semantics.
Furthermore, having a non-single channel prior to the linear projection to query, key and value matrices is architecturally natural for a transformer-based architecture.

\subsection{Match-to-match attention}

\noindent\textbf{Attention layer.}
We flatten the 4D correlation map to behave as the input sequence for the transformer module, \ie, $\mathbbm{R}^{L \times H \times W \times H \times W} \rightarrow \mathbbm{R}^{L \times HWHW}$, considering the match at each spatial position as an element for attention.
We then linearly embed the channel dimension of our flattened correlation map, \ie, $\mathbf{X} = \mathbf{C}^\top\mathbf{W}_{\text{in}}$, where $\mathbf{C}$ refers to the correlation map, $\mathbf{W}_{\text{in}} \in \mathbb{R}^{L \times D_{\text{in}}}$ is the linear transformation matrix, and $\mathbf{X} \in \mathbb{R}^{HWHW \times D_{\text{in}}}$ is the input to the subsequent attention blocks.
However, the quadratic complexity of conventional self-attention in transformers poses an infeasible computation overhead in our setting, as a flattened 4D tensor results in a significantly long 1D tensor.

Inspired by Fastformer~\cite{wu2021fastformer}, we aim to alleviate this bottleneck through the use of \emph{additive} attention to effectively model long-range match-to-match interactions; instead of computing a quartic attention map (with respect to the spatial size of feature maps) which encodes all possible interactions between candidate matches $\mathbf{Q}\mathbf{K{^\top}} \in \mathbb{R}^{T \times T}$ where $T=HWHW$, we form a compact representation of query-key interactions $\mathbf{H} \in \mathbb{R}^{T \times D_h}$ via additive attention which computes interactions between a global query representation and every key vector:
\begin{align}
    \label{eq:additive_attention_P}
    \mathbf{H}^{(h)}_{i,:} = \mathbf{K}^{(h)}_{i,:} \odot \sum_{j=1}^{T}\mathbf{Q}^{(h)}_{j,:} \sigma(\tau\textbf{w}_{\mathrm{q}}\textbf{Q}^{(h)\top})_{j},
\end{align}
where $\mathbf{w}_{\mathrm{q}} \in \mathbb{R}^{D_h}$ learns to transform the query vectors into a global vector.
A similar additive attention mechanism summarizes the context-aware key representations $\mathbf{H}$ with a linear projection $\mathbf{w}_{\mathrm{k}} \in \mathbb{R}^{D_h}$ to model its interaction with value vectors as follows:
\begin{align}
    \label{eq:fastformer}
    \text{SA}^{(h)}_{\text{TM}}(\mathbf{X})_{i,:} = \mathbf{V}^{(h)}_{i,:} \odot \sum_{j=1}^{T} \mathbf{H}^{(h)}_{j,:} \sigma(\tau\textbf{w}_{\mathrm{k}} \mathbf{H}^{(h)\top})_{j},
\end{align}
with the assumption of $D_h = D_v$.
The output is transformed by an MLP followed by residual connection with $\mathbf{Q}$.
Our proposed match-to-match attention layer reduces the time and memory complexity down to linear with respect to the input length: $\mathcal{O}(T^2D_h) \xrightarrow{} \mathcal{O}(TD_h)$. 

Finally, to ensure that our attention layer can attend to parts of the flattened correlation map differently, we formulate our multi-head self-attention layer as follows:
\begin{align}
    \label{eq:MHSA_tmatcher}
    \text{MHSA}_{\text{TM}}(\mathbf{X}) = \underset{h \in [N_h]}{\text{concat}}\big[\text{SA}_{\text{TM}}^{(h)}(\mathbf{X})\big]\textbf{W}_{\mathrm{O}} + \mathbf{b}_{\mathrm{O}}.
\end{align}
where a linear transformation layer transforms the concatenated outputs of the multiple self-attention layers.
We use the pre-LN approach, where the layer normalization is placed inside the residual blocks of the attention layers.

\noindent\textbf{4D rotary positional embedding.}
In transformer-based networks, positional embedding models the dependency between elements at different positions in the sequence.
While relative positional embedding has shown to outperform absolute positional embedding in modelling relation-aware interactions, it is not applicable to linear-complexity transformers as they do not explicitly compute the quadratic-complexity attention matrix.
To this end, we employ rotary positional embedding (RoPE)~\cite{su2021roformer} and extend it to be applicable on our 4D correlation map input.

RoPE aims to make the interaction of query and key (inner product for vanilla transformers) encode the position information only in the relative form.
Their proposed attention matrix computation with RoPE in vanilla quadratic-complexity transformers can be formulated as follows:
\begin{align}
\label{eq:rpe_attention}
    \mathbf{Q}^{(h)}_{m,:}\mathbf{K}_{n,:}^{(h)\top} &= (\mathbf{X}_{m,:}\mathbf{W}^{(h)}_\mathrm{Q}\mathbf{R}_{(\Theta,m)})(\mathbf{X}_{n,:}\mathbf{W}^{(h)}_\mathrm{K}\mathbf{R}_{(\Theta,n)})^\top \\
    &= \mathbf{X}_{m,:}\mathbf{W}^{(h)}_\mathrm{Q}\mathbf{R}_{(\Theta,n-m)}\mathbf{W}^{(h)\top}_\mathrm{K}\mathbf{X}^{\top}_{n,:},
\end{align}
where $\mathbf{R}_{(\Theta,*)} \in \mathbb{R}^{D_h \times D_h}$ is the rotary matrix which is for rotating the key or query vectors by amount of angle in multiples of their position indices to incorporate relative positional embedding. 
We guide the readers to the supplementary for detailed explanations.

RoPE can be applied to linear-complexity transformers as well~\cite{su2021roformer}.
In our work, we achieve this by by using \Eq{additive_attention_P} to calculate global context-aware query-key interactions, but with $\mathbf{K} = \mathbf{X}\mathbf{W}_\textrm{K}\mathbf{R}_{(\Theta,*)}$ and $\mathbf{Q} = \mathbf{X}\mathbf{W}_\textrm{Q}\mathbf{R}_{(\Theta,*)}$.

\noindent\textbf{Single-channel refined correlation computation.}
In a nutshell, our match-to-match module takes as input a noisy 4D correlation map to refine it using match-to-match interactions, outputting a refined correlation map for robust image matching.
This process is repeated $N$ times, providing a tensor in $\mathbbm{R}^{L \times HWHW}$.
The output from the final match-to-match attention module is linearly projected to a single channel dimension, and is reshaped back to 4D correlation map \ie $\mathbbm{R}^{L \times HWHW} \rightarrow \mathbbm{R}^{H \times W \times H \times W}$, for reliable keypoint transfer.
For precise transfer, we perform a 4-dimensional upsampling function on the 4D correlation map, and denote the tensor as $\mathbf{C}^{\text{out}} \in \mathbb{R}^{\bar{H} \times \bar{W} \times \bar{H} \times \bar{W}}$ where $\bar{H} = 2H$ and $\bar{W} = 2W$ which corresponds to $\frac{1}{8}$ the size of the original image.
We illustrate the outline of our match-to-match attention module in \Fig{fastformer_detail}.

\subsection{Flow field formation}

The output correlation tensor $\mathbf{C}^{\text{out}}$ can be transformed into a dense flow field by applying kernel soft-argmax~\cite{lee2019sfnet}. 
We normalize the raw correlation outputs using softmax:
\begin{align}
    \mathbf{C}^{\text{norm}} = \frac{\text{exp}(\mathbf{G}_{kl}^\mathbf{p}\mathbf{C}^{\text{out}}_{ijkl})}{\sum_{(k',l')\in \bar{H} \times \bar{W}}\text{exp}(\mathbf{G}_{k'l'}^\mathbf{p}\mathbf{C}^{\text{out}}_{ijk'l'})},
\end{align}
where $\mathbf{G}^\mathbf{p} \in \mathbb{R}^{\bar{H} \times \bar{W}}$ is a 2-dimensional Gaussian kernel centered on $\mathbf{p} = \text{arg max}_{k,l} \mathbf{C}^{\text{out}}_{i,j,k,l}$, which is applied to smooth the potentially irregular correlation values.
The normalized correlation tensor $\mathbf{C}^{\text{norm}}$ encodes a set of probability simplexes, which we use to transfer all the coordinates on the dense regular grid $\mathbf{P} \in \mathbb{R}^{\bar{H} \times \bar{W} \times 2}$ of source image $I$ to obtain their corresponding coordinates $\mathbf{\hat{P}'} \in \mathbb{R}^{\bar{H} \times \bar{W} \times 2}$ on target image $\hat{I}$: $\hat{\mathbf{R}}'_{i,j} = \sum_{(k,l)\in \bar{H} \times \bar{W}} \mathbf{C}^{\text{norm}}_{i,j,k,l}\mathbf{P}_{k,l}$.
We then can construct a dense flow field at sub-pixel level using the set of estimated matches $(\mathbf{P},\mathbf{\hat{P}}')$.

\input{tables/semantic_matching}

\subsection{Training objective}
We assume that we are given a set of ground-truth coordinate pairs $\mathcal{M} = \{(\mathbf{k}_m, \hat{\mathbf{k}}_m)\}_{m=1}^M$ for each training image pair, where $M$ is the number of annotated keypoint matches.
We carry out keypoint transfer from the source to the target keypoints using the constructed dense flow field.
For a given keypoint $\mathbf{k} = (x_k,y_k)$, we define a soft sampler $\textbf{W}^{(k)} \in \mathbb{R}^{\bar{H} \times \bar{W}}$:
\begin{align}
\mathbf{W}_{ij}^{(k)} = \frac{\text{max}(0, \tau - \sqrt{(x_k - j)^2 + (y_k - i)^2})}{\sum_{i'j'} \text{max}(0, \tau - \sqrt{(x_k - j')^2 + (y_k - i')^2})},
\end{align}
where $\tau$ is a distance threshold, and $\sum_{ij} \mathbf{W}_{ij}^{(k)} = 1$.
It can be seen that the soft sampler effectively samples each transferred keypoint $\mathbf{\hat{P}}'_{ij}$ by assigning weights inversely proportional to the distance to $\mathbf{k}$.
Using this soft sampler, we assign a match to the keypoint $\mathbf{k}$ as $\hat{\mathbf{k}}' = \sum_{(i,j) \in \bar{H} \times \bar{W}}\mathbf{\hat{P}}'_{ij:}\mathbf{W}_{ij}^{(k)}$, being able to achieve up to sub-pixel-wise accurate keypoint matches.
By applying this keypoint transfer method on the source keypoints, we obtain the predicted keypoint pairs on image $\hat{I}: \{(\textbf{k}_m, \hat{\textbf{k}}'_m)\}_{m=1}^M$ by assigning a match $\hat{\textbf{k}'}_m$ to each keypoint $\textbf{k}_m$ in the source image.
We formulate our training objective to minimize the average Euclidean distance between the predicted target keypoints and the ground-truth target keypoints as follows:
\begin{align}
\mathcal{L} = \frac{1}{M}\sum_{m=1}^M \|\hat{\mathbf{k}}_m - \hat{\mathbf{k}}'_m \|_{2}^{2}.
\end{align}

%% file: tables/semantic_matching.tex
\begin{table*}
\centering
\scalebox{0.95}{
\begin{tabular*}{\textwidth}{l@{\extracolsep{\fill}}ccccccccc}
                \toprule
                \multirow{3}{*}{Method} & \multicolumn{2}{c}{SPair-71k} & \multicolumn{2}{c}{PF-PASCAL} & \multicolumn{2}{c}{PF-WILLOW} & \multirow{3}{*}{\shortstack{time\\(\emph{ms})}} & \multirow{3}{*}{\shortstack{memory\\(GB)}} &\multirow{3}{*}{\shortstack{FLOPs\\(G)}}\\
                
                & \multicolumn{2}{c}{@$\alpha_{\text{bbox}}$} & \multicolumn{2}{c}{@$\alpha_{\text{img}}$} & @$\alpha_{\text{bbox-kp}}$ & @$\alpha_{\text{bbox}}$ \\ 
                 
                 & 0.1 (F) & 0.1 (T) & 0.05 (F) & 0.1 (F) & 0.1 (T) & 0.1 (T)\\
                 
                 \midrule
                 
                 NC-Net~\cite{rocco2018neighbourhood} & 20.1 & 26.4 & {54.3} & 78.9 & 67.0 & - & 222 & 1.2 & 44.9 \\
                 
                 DCC-Net~\cite{huang2019dynamic}  & - & 26.7 & {55.6} & {82.3} & 73.8 & - & 567 & 2.7 & 47.1 \\
                 
                 DHPF~\cite{min2020dhpf}  & 27.7 & 28.5 & {56.1} & {82.1} & 74.1 & \textbf{80.2} & 58 & 1.6 & 2.0\\
                 
                 PMD~\cite{li2021pmdnet} &  26.5 & - & - & 81.2 & 74.7 & - & - & - & -\\
                 
                 \midrule

                 UCN~\cite{choy2016universal}          & - & 17.7 & - & 75.1 & - & - & - & - & -\\
                 
                 HPF~\cite{min2019hyperpixel}      &  28.2 & - & 60.1 & 84.8 & 74.4 & - & 63 & - & -\\
                 
                 SCOT~\cite{liu2020semantic} & 35.6 & - & 63.1 & 85.4 & 76.0 & - & 151 & 4.6 & 6.2 \\
                 
                 SCNet~\cite{han2017scnet}           & - & - & 36.2 & 72.2 & - & 70.4 & $>$1000 & - & -\\
                 
                 DHPF~\cite{min2020dhpf}           & {37.3} & 27.4  & {75.7} & {90.7} & 71.0 & 77.6 & 58 & 1.6 & 2.0\\
                 
                 DHPF$\dagger$ ~\cite{min2020dhpf}         & 39.4 & -  & - & - & - & - & 58 & 1.6 & 2.0\\
                 
                 NC-Net$^\textrm{\textbf{*}}$~\cite{rocco2018neighbourhood} & - & - & - & 81.9 & - & -& 222 & 1.2 & 44.9\\
                 
                 DCC-Net$^\textrm{\textbf{*}}$~\cite{huang2019dynamic} &  - & - & - & 83.7 & - & - & 567 & 2.7 & 47.1\\
                 
                 ANC-Net~\cite{li2020correspondence} &  - & 28.7 & - & 86.1 & - & - & 216 & 0.9 & 44.9\\
                 
                 PMD~\cite{li2021pmdnet} &  37.4 & - & - & 90.7 & 75.6 & - & - & - & -\\
                 
                 CHMNet~\cite{min2021chm}   & {46.3} & \underline{30.1} & 80.1 & {91.6} & 69.6 & \underline{79.4} & 54 & 1.6 & 19.6\\
                 
                 PMNC~\cite{lee2020pmnc} &  \underline{50.4} & - & \textbf{82.4} & 90.6 & - & - & - & - & -\\
                 
                 MMNet~\cite{zhao2021multi} &  40.9 & - & 77.6 & 89.1 & - & - & 86 & - & -\\
                 
                 CATs~\cite{cho2021semantic}   & 43.5 & - & - & - & - & - & 45 & 1.6 & 28.4\\
                 
                 CATs$\dagger$~\cite{cho2021semantic}   & 49.9 & 27.1 & 75.4 & \textbf{92.6} & 69.0 & 79.2 & 45 & 1.6 & 28.4\\
                 
                 \midrule
                 \midrule
                 
                 TransforMatcher (ours)   & 50.2 & \textbf{30.5} & 78.9 & 90.5 & 66.7 & 75.1 & 54 & 1.6 & 33.5 \\
                 TransforMatcher$\dagger$ (ours)   & \textbf{53.7} & \underline{30.1} & \underline{80.8} & \underline{91.8} & 65.3 & 76.0 & 54 & 1.6 & 33.5  \\
                 
                 \bottomrule
        \end{tabular*}
        }
        \vspace{-1.0mm}
        \caption{\textbf{Performance on standard benchmarks of semantic matching.} Higher PCK is better.
            All the results reported in the table uses pretrained ResNet-101 model as the feature extractor.
            Methods in the first group were trained with weak supervision (image pair annotations), while those in the second group were trained with strong supervision (sparse keypoint match annotations).
            Models with $\textrm{\textbf{*}}$ are retrained using keypoint annotations from ANC-Net~\cite{li2020correspondence}.
            $\dagger$ indicates the use of data augmentation during training.
            Numbers in bold indicate the best performance, followed by the underlined numbers.
            Some results are from~\cite{min2021chm}.
        }
        \label{tbl:main_table}
        \vspace{-1.0mm}
\end{table*}

%% file: sections/5_experiments.tex
\input{tables/ablation_posembedding}

\section{Experiments}

We evaluate our method on the semantic correspondence task, which aims to match semantically similar parts between images of the same category but different instances.

\smallbreak
\noindent \textbf{Datasets.}
We report our results on standard benchmark datasets of semantic correspondence: SPair-71k~\cite{min2019spair}, PF-PASCAL~\cite{ham2018proposal}, and PF-WILLOW~\cite{ham2016proposal}.
The SPair-71k dataset has diverse variations in viewpoint and scale, with 53,340 / 5,384 / 12,234 image pairs for training, validation, and testing, respectively.
The PF-PASCAL and PF-WILLOW datasets are taken from four categories of the PASCAL VOC dataset, having small viewpoint and scale variations.
The PF-PASCAL dataset contains 2,940 / 308 / 299 image pairs for training, validation and testing, respectively.
The PF-WILLOW dataset contains 900 image pairs for testing only.
The SPair-71k dataset is significantly larger than the other two datasets, and has more accurate and richer annotations regarding different levels of difficulty in occlusion, truncation, viewpoint and illumination.
Being the most challenging dataset, the results on SPair-71k are less saturated in comparison.

\smallbreak
\noindent \textbf{Implementation details.}
Following recent methods~\cite{min2021chm, cho2021semantic}, we employ the ResNet-101 model pre-trained on the ImageNet classification task~\cite{krizhevsky2012imagenet} as the feature extraction network.
Note that the \texttt{conv4\_x} and \texttt{conv5\_x} layers in ResNet-101 have 23 and 3 bottleneck layers respectively, from which we extract feature maps to compute 26 layer-wise correlations maps for each image pair.
We set the spatial size of the input image to $240 \times 240$, resulting in $H = W = 15$ for feature maps used for correlation computation, and $\bar{H} = \bar{W} = 30$.
Each of our match-to-match attention layers have 8 heads for multi-head self attention ($N_h=8$), with head dimension of 4 ($D_h=D_v=4$). 
The overall pipeline of our method is implemented using PyTorch~\cite{NEURIPS2019_9015}, and is optimized using the Adam~\cite{kingma2015adam} optimizer with a constant learning rate of 1e-3.
We finetune the feature extractor network at a lower learning rate of 1e-5.

\smallbreak
\noindent \textbf{Evaluation metric.}
We use the percentage of correct keypoints (PCK) for evaluation, which is the standard evaluation metric for category-level matching.
Given a pair of ground-truth and predicted target keypoints $\mathcal{K}=\{(\hat{\mathbf{k}}_m, \hat{\mathbf{k}}'_m)\}_{m=1}^M$, PCK is measured by: 
\begin{align}
\text{PCK}(\mathcal{K}) = \frac{1}{M}\sum_{m=1}^M\mathbbm{1}[\|\mathbf{\hat{k}}_m - \hat{\mathbf{k}}'_m\| \leq \alpha_{\tau}\cdot\text{max}(w_{\tau},h_{\tau})],
\end{align}
where $w_{\tau}$ and $h_{\tau}$ are the width and height of either the entire image or the object bounding box, \ie, $\tau \in \{\text{img, bbox-kp, bbox}\}$, and $\alpha_{\tau}$ is a tolerance factor.

\subsection{Results and analysis.}
\input{figures/qualitative}

For the SPair-71k dataset, we evaluate two versions for our model: a finetuned model (F) trained on SPair-71k, and a transferred model (T) trained on PF-PASCAL. 
On the PF-PASCAL and PF-WILLOW datasets, we follow the common evaluation protocol to train our network on the training split of PF-PASCAL and evaluate on the test splits of PF-PASCAL and PF-WILLOW.
The quantitative results are illustrated in \Tbl{main_table}.
Previous methods have been using two different schemes, \eg, $\tau \in \{\text{bbox-kp, bbox}\}$, when computing the threshold for PF-WILLOW~\cite{min2021cpchm}, so we report our results using both thresholds.

We show that TransforMatcher finetuned on SPair-71k sets a new state of the art.
A notable observation is that TransforMatcher finetuned on SPair-71k \emph{without} data augmentation outperforms CATs~\cite{cho2021semantic} trained \emph{with} augmentation, proving the efficacy of our 4D match-to-match attention and multi-level correlation score features.
Using data augmentations leads to improved PCK on both SPair-71k and PF-PASCAL datasets, but transformer-based models benefit more from augmentations as seen from the lower PCK increase in DHPF~\cite{cho2021semantic}.
It is interesting that TransforMatcher trained without data augmentations transfer slightly better to SPair-71k and PF-WILLOW datasets than our model trained with data augmentations, albeit its lower PCK performance on PF-PASCAL.
This potentially hints that while data augmentations do help TransforMatcher to learn better, it overfits more to the training data domain.
TransforMatcher also exhibits state-of-the-art performance when transferred to the SPair-71k dataset, while being comparable on the PF-PASCAL dataset.
However, TransforMatcher shows substandard results when transferred to the PF-WILLOW dataset, unlike the SPair-71k dataset.
This evidences that the match-to-match interactions learned from the PF-PASCAL dataset is better transferable to the SPair-71k dataset, but is not as effective on the PF-WILLOW dataset.
\Fig{qualitative} visualizes example qualitative results on SPair-71K using our model.

\subsection{Ablation study and analysis}

\smallbreak
\noindent \textbf{Effect of data augmentation during training.}
CATs\cite{cho2021semantic} found that using data augmentation for category-level matching model is beneficial, especially for data-hungry transformer-based architectures.
We study the effect of applying data augmentation to our model as well, following the schemes used in CATs. 
The results in \Tbl{main_ablation} show that using data augmentation indeed gives consistent improvements to the performance of our model.

\smallbreak
\noindent \textbf{Analysis on positional embedding.}
We investigate the effect of positional embedding used in our pipeline.
As conventional relative positional embedding requires an explicit computation of the attention matrix, is not applicable to our transformer architecture with the linear-complexity additive attention.
On the other hand, rotary positional embeddings can be seamlessly applied to our model as an alternative method to model relative positional embedding.
The results in \Tbl{main_ablation} show that using rotary positional embedding results in significant gains over absolute positional embedding, especially on the more challenging SPair-71k dataset.

\smallbreak
\noindent \textbf{Analysis on efficient transformer architecture.} 
We try replacing our match-to-match attention architecture with other efficient transformer designs~\cite{wang2020linformer, choromanski2020rethinking}, and also the vanilla transformer~\cite{vaswani2017attention} design to compare the performances.
We use absolute learnable positional embedding in this experiment.
The results in \Tbl{small_ablation} show that the additive attention architecture shows the most favorable results, with similarly high performance as Performer but with lower latency.
We found that the Linformer architecture~\cite{wang2020linformer} failed to train, which we conjecture is due to the low head dimension of our network, and the reliance of Linformer on kernel approximations which could lead to inaccurate interactions between the position-sensitive matches.
Training with vanilla Transformers was infeasible due to its large memory demands of the pair-wise attention matrices.

\begin{figure}[t] 
    \begin{center}
        \includegraphics[width=0.99\linewidth]{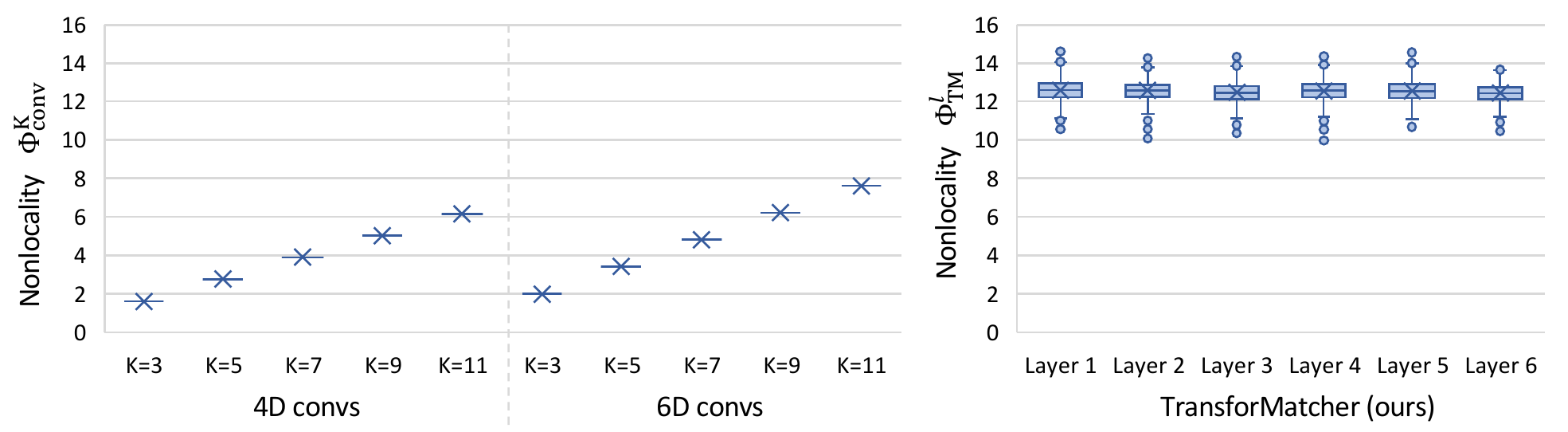}
    \end{center}
    \vspace{-6.0mm}
      \caption{\textbf{Nonlocality distributions of high-dim. conv kernels (left) and TransforMatcher's attention layers (right)}.}
    \vspace{-3.0mm}
\label{fig:boxplot}
\end{figure}

\begin{figure}[t] 
    \begin{center}
        \includegraphics[width=0.99\linewidth]{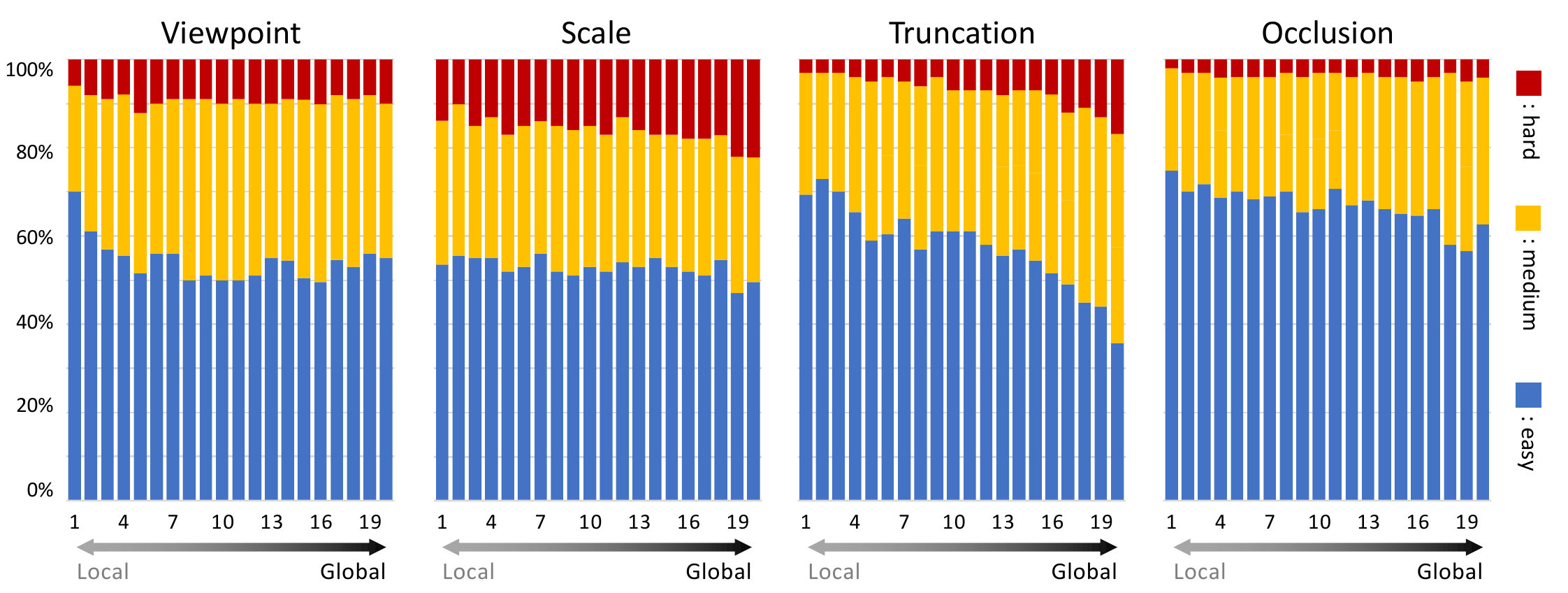}
    \end{center}
    \vspace{-7.0mm}
      \caption{\textbf{Proportion of image pair difficulty w.r.t. nonlocality.}}
    \vspace{-5.0mm}
\label{fig:stats}
\end{figure}

\smallbreak
\noindent \textbf{Analysis on nonlocality of match-to-match attention.}
For an in-depth analysis, we investigate how nonlocally our match-to-match attention layers operate in comparison to convolutional counterparts~\cite{min2021chm, rocco2018neighbourhood}.
We define the measure of nonlocality of an MHSA at layer $l$ as the average of interactions between attention scores and relative offsets:
\begin{small}
\begin{align}
    \label{eqn:nonlocality}
    \Phi^{l} = \frac{1}{Z} \sum_{h \in [N_h]} \sum_{(\mathbf{q}, \mathbf{k}) \in \mathcal{X} \times \mathcal{X}} \mathbf{A}_{\mathbf{q}, \mathbf{k}}^{(h)} \|\mathbf{q}-\mathbf{k}\|^{2}, 
\end{align}
\end{small}
where $Z$ is normalization constant and $\mathcal{X}$ is a set of spatial positions in $\mathbf{C}$.
Figure~\ref{fig:boxplot} plots distributions of nonlocality values for high-dim convolutional layers and MHSA layers in TransforMatcher; convolutional layers layers {\em statically} transforms with {\em fixed}, {\em local} receptive fields ($\Phi^{K}_{\text{conv}} < 8$) regardless of input contents.
In contrast, Transformatcher layers can \textit{dynamically} transform input contents by \textit{adaptively} deciding regions of attention for effective transformation with \textit{global} receptive fields ($\Phi^{l}_{\text{TM}} \approx 12.5$).
To verify the benefits of dynamic global match-to-match attention, we measure sample-wise nonlocality ($\Phi=\sum_{l=1}^{L}\Phi^{l}$) for each test image pair in the SPair-71k, assort them into 20 groups with increasing nonlocality, and visualize the proportion of the difficulty levels for each group in Fig.~\ref{fig:stats}.
For all difficulty types, the proportion of hard/medium samples increase with increasing nonlocality. 
This trend is especially visible in types of truncation/occlusion; our model attends larger contexts to better perceive truncated/occluded parts.
We guide the readers to the supplementary material for the implementation details of this analysis, together with additional analyses and qualitative results of TransforMatcher. 

%% file: tables/ablation_posembedding.tex
\begin{table*}[!t]
    \parbox{.49\linewidth}{
    \centering
        \scalebox{0.9}{
       \begin{tabular}{c@{\extracolsep{\fill}}ccccc}
                \toprule
                \multirow{3}{*}{Augmentation} & \multirow{3}{*}{\shortstack{Positional\\Embedding}} &  \multicolumn{2}{c}{SPair-71k} & \multicolumn{2}{c}{PF-PASCAL} \\
                
                                              &                                 &  \multicolumn{2}{c}{@$\alpha_{\text{bbox}}$} & \multicolumn{2}{c}{@$\alpha_{\text{img}}$} \\ 
                 
                                              &                                 & 0.05 & 0.1 & 0.05 & 0.1\\
                 \midrule
                 
                                       &  Absolute~\cite{ott2019fairseq}                    & 29.9 & 48.7 & 74.5 & 89.4 \\
                \checkmark                       &  Absolute~\cite{ott2019fairseq}                   & 26.6 & 48.9 & \underline{79.4} & \textbf{91.8}\\
                                       & Rotary~\cite{su2021roformer}                      & \underline{30.5} & \underline{50.2} & 78.9 & 90.4 \\
                \midrule
                 \checkmark                       & Rotary~\cite{su2021roformer}                      & \textbf{32.4} & \textbf{53.7} & \textbf{80.8} & \textbf{91.8} \\
                 \bottomrule
        \end{tabular}
        }
        \vspace{-2.0mm}
        \caption{\textbf{Ablation on augmentation and positional embedding.} 
        The results show that using data augmentation and rotary positional embedding gives the best results.}
        \label{tbl:main_ablation}
        \vspace{-3.0mm}
    }
    \hfill
    \parbox{.49\linewidth}{
    \centering
        \scalebox{0.9}{
        \begin{tabular}{l@{\extracolsep{\fill}}ccccc}
                \toprule
                \multirow{3}{*}{Architecture} & \multicolumn{2}{c}{SPair-71k} & \multirow{3}{*}{\shortstack{time\\(\emph{ms})}} & \multirow{3}{*}{\shortstack{memory\\(GB)}} &\multirow{3}{*}{\shortstack{FLOPs\\(G)}}\\
                                  & \multicolumn{2}{c}{@$\alpha_{\text{bbox}}$}\\ 
                                  &  0.05 & 0.1  \\
                \midrule
                Transformer~\cite{vaswani2017attention} & - & - & \multicolumn{3}{c}{Out-Of-Memory} \\
                Linformer~\cite{wang2020linformer}           &  0.34 & 1.3 & 36 & 1.7 & 33.4\\
                Performer~\cite{choromanski2020rethinking}  &  \textbf{28.2} & 48.8 & 88 & 1.6 & 35.9\\
                \midrule
                Additive Attn.           &  26.6 & \textbf{48.9} & 54 & 1.6 & 33.5 \\
                \bottomrule
                
        \end{tabular}
        }
        \vspace{-2.0mm}
        \caption{\textbf{Results of different transformer architectures.} 
        Vanilla transformer could not be evaluated within memory capabilities.
        Additive attention yields the most favorable results.
        }
        \label{tbl:small_ablation}
        \vspace{-3.0mm}
    }
\end{table*}

%% file: figures/qualitative.tex
\begin{figure}[ht]
  \centering
    \begin{center}
    \includegraphics[width=0.45\textwidth]{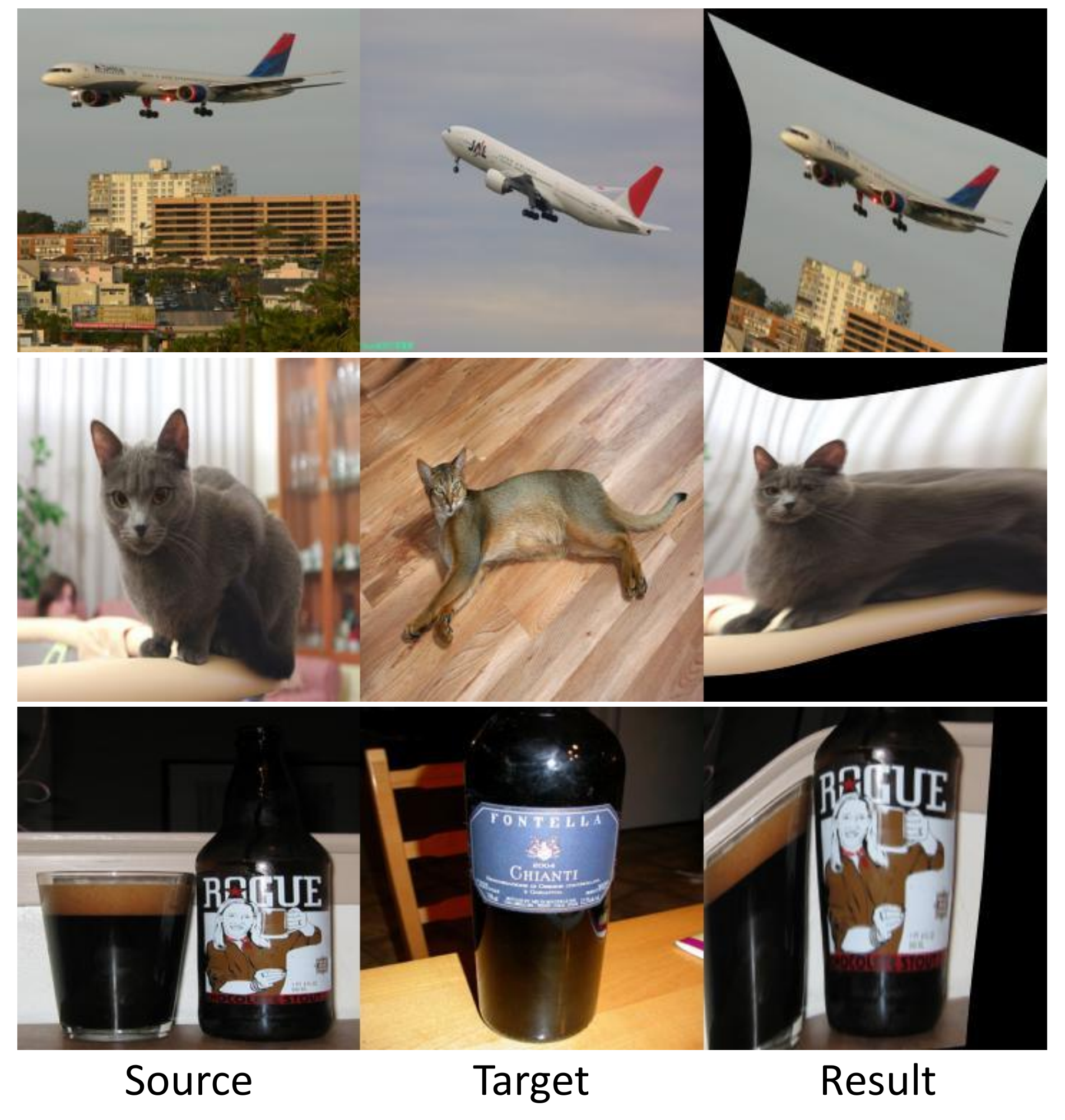}
    \end{center}
  \vspace{-4.0mm}
  \caption{\textbf{Sample results on SPair-71k.} Source images are TPS-transformed~\cite{donato2002approximate} to target images using predicted correspondences.
  }
  \label{fig:qualitative}
  \vspace{-5.0mm}
\end{figure}

%% file: sections/6_conclusion.tex
\section{Conclusion}
In this paper, we have proposed the TransforMatcher, an effective semantic matching learner.
Our principal contribution is the match-to-match attention mechanism, which is, to the best of our knowledge, the first attempt to directly process a 4D input, \ie, correlation map, with every spatial entry (match) as an element for attention using a transformer-based network with {\em global} receptive fields. 
This has been a challenging pursuit due to the quadratic complexity of vanilla transformers in modeling global-range interactions, which was addressed by additive attention with linear complexity.
We further propose to treat multi-level correlation scores as features to better exploit the richer semantics in different levels of feature maps.
The proposed model outperforms state of the arts on the SPair-71k dataset, while performing on par with the SOTA methods on the PF-PASCAL dataset. 
While the memory usage of TransforMatcher increases quadratically with respect to the number of pixels as in other dense matching methods, we anticipate this work will motivate the use of transformers with high-dimensional inputs in other domains.

%% file: 02_supp.tex
\setcounter{section}{0}
\setcounter{table}{0}
\setcounter{figure}{0}
\renewcommand{\thesection}{\Alph{section}}
\renewcommand\thefigure{A\arabic{figure}}
\renewcommand{\thetable}{A\arabic{table}}

\title{TransforMatcher: Match-to-Match Attention for Semantic Correspondence \\ 
{\it ----- Supplementary Material -----}}

\maketitle
\input{tables/supp_classwise}

In this supplementary material, we provide additional details, results and analyses of our proposed TransforMatcher pipeline.

\section{Rotary positional embedding details}

To keep the paper self-contained, we briefly explain on the formulation of rotary positional embedding (RoPE)~\cite{su2021roformer}. 
The aim of RoPE is to find an encoding mechanism $f_{\{ q,k\}}$ such that the inner product, $g$, of query $q_m$ and key $k_n$ of embeddings $\mathbf{x}_m, \mathbf{x}_n \in \mathbb{R}^{d}$ encodes position information only in the relative form as follows:
\begin{align}
    \label{eq:rope_formulation}
    \langle f_q(\mathbf{x}_m,m), f_k(\mathbf{x}_n, n) \rangle = g(\mathbf{x}_m, \mathbf{x}_n, m-n),
\end{align}
where $m-n$ denotes the relative position between the embeddings. Starting from a simple case with dimension $d=2$, RoPE exploits the geometric properties of vectors on 2D plane and its complex form to prove that a solution to \Eq{rope_formulation} is:
\begin{align}
    \label{eq:rope_2Dcase}
    f_q(x_m,m) &= (\mathbf{W}_q\mathbf{x}_m)e^{im\theta}, \\
    f_k(x_n,n) &= (\mathbf{W}_k\mathbf{x}_n)e^{in\theta}, \\
    g(x_m, x_n, m-n) &= \textrm{Re}[(\textbf{W}_q\mathbf{x}_m)(\textbf{W}_k\mathbf{x}_n)\textrm{*}e^{i(m-n)\theta}],
\end{align}
where $\textrm{Re}[\cdot]$ is the real part of a complex number, $(\textbf{W}_k\mathbf{x}_n)\textrm{*}$ is the conjugate complex number of $(\textbf{W}_k\mathbf{x}_n)$, and $\theta \in \mathbb{R}$ is a predefined non-zero constant. 
Writing $f_{\{ q,k\}}$ in the form of matrix multiplication gives:
\begin{align}
    \label{eq:rope_2Dcase_matmul}
    &f_{\{q,k\}}(\mathbf{x}_m,m) = \nonumber \\
    &
    \begin{pmatrix}
    \textrm{cos }m\theta & -\textrm{sin }m\theta\\
    \textrm{sin }m\theta & \textrm{cos }m\theta 
    \end{pmatrix}
    \begin{pmatrix}
    \textbf{W}^{(11)}_{\{q,k\}} & \textbf{W}^{(12)}_{\{q,k\}} \\
    \textbf{W}^{(12)}_{\{q,k\}} & \textbf{W}^{(22)}_{\{q,k\}}
    \end{pmatrix}
    \begin{pmatrix}
     x_m^{(1)}\\
     x_m^{(2)}
    \end{pmatrix},
\end{align}
where $[x_{m}^{(1)}, x_{m}^{(2)}]^{\top} = \mathbf{x}_m$ given $d=2$.
Henceforth, to incorporate relative positional embedding, we can simply rotate the key/query embedding by amount of angle in multiples of its position index.
The above formulation can be generalized to any even dimension $d$, by dividing the $d$-dimension space to $\frac{d}{2}$ sub-spaces, which are combined using the linearity of inner product.
We refer the readers to the original paper~\cite{su2021roformer} for full details. 

\section{Additional results and analyses}

\noindent \textbf{Category-wise PCK results.}
We show the category-wise PCK results of our model on the SPair-71k dataset~\cite{min2019spair} in comparison to existing methods in \Tbl{supp_classwise}.
It can be seen that TransforMatcher achieves the highest PCK overall, and the highest PCK in the majority of categories. 
An interesting observation is that while CATs~\cite{cho2021semantic} trained with augmentation shows consistently improved results compared to using no augmentation, TransforMatcher trained without augmentation often shows higher PCK values compared to TransforMatcher trained with augmentation. 
We conjecture this is because CATs also processes the actual 2D feature maps of source and target images together with the 4D correlation map using transformers, while TransforMatcher relies only on the 4D correlation map to find correspondences.
An important takeaway is that is that while leveraging data augmentation provides more accurate semantic correspondences overall, it may have adverse effects on certain categories depending on the network architecture.

\begin{figure}[t]
    \centering    
    \begin{center}
        \includegraphics[width=0.85\linewidth]{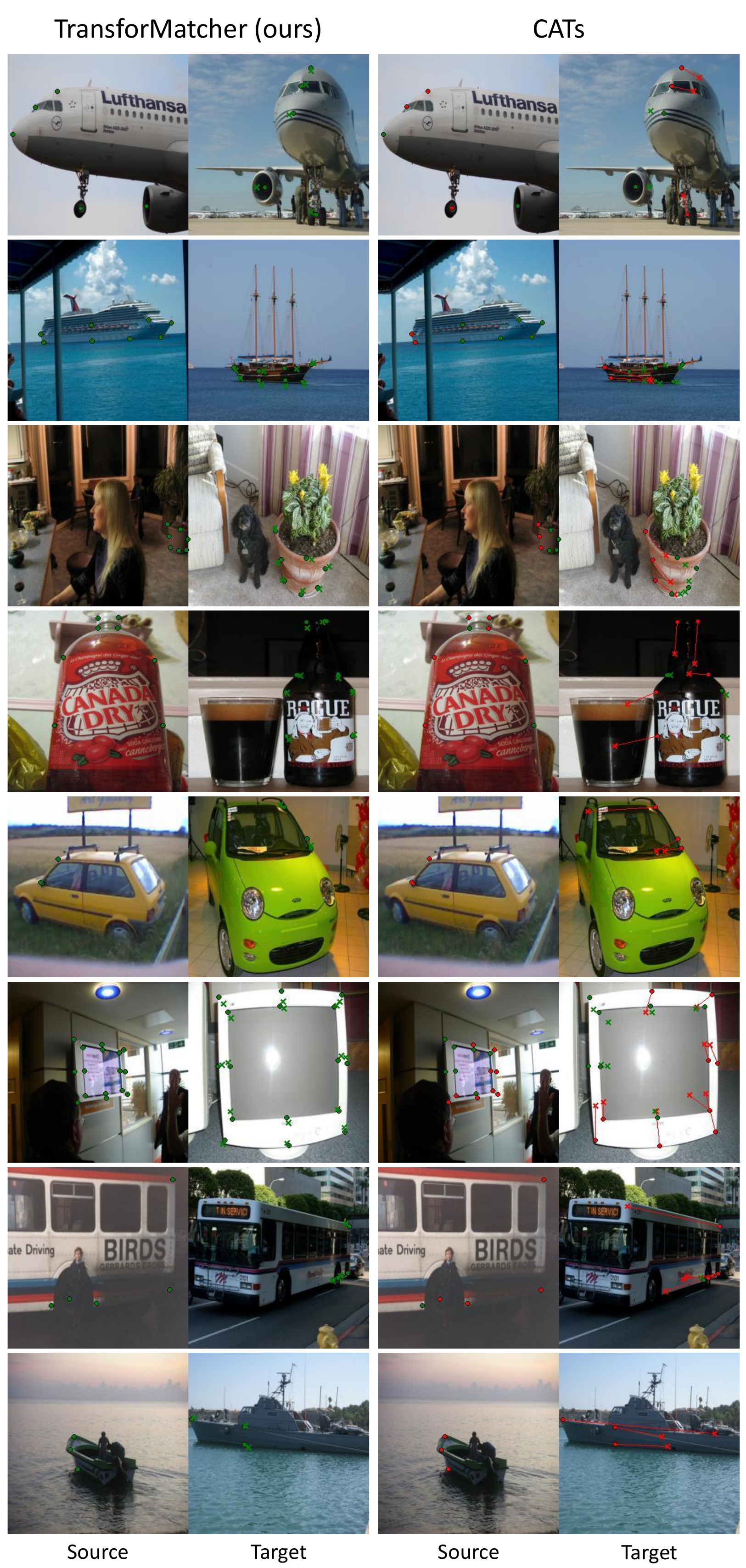}
    \end{center}
    \vspace{-5.0mm}
      \caption{Qualitative comparison between the proposed TransforMatcher (left) and CATs~\cite{cho2021semantic} (right). We show keypoints in circles and predictions in crosses with a line that depicts matching error. Best viewed in electronic forms.}
    \vspace{-5.0mm}
\label{fig:vis_comp}
\end{figure}

\input{tables/supp_multichannel}
\noindent \textbf{Ablation on correlation map channel dimension.}
We stated in the main paper that we construct a multi-channel correlation map as it is architecturally natural, and to exploit the richer semantics in different levels of feature maps.
We conduct an experiment to compare the results of TransforMatcher when using a single-channel correlation map instead of a multi-channel correlation map.
For fairness, we use the same bottleneck layers of \texttt{conv4\_x} and \texttt{conv5\_x}, and construct a single-channel correlation map by either (1) concatenating the multi-layer features along the channel dimension prior to correlation computation(Single$_{\textrm{concat}}$), or (2)  taking the mean of the multi-channel correlation map(Single$_{\textrm{mean}}$).
\Tbl{supp_channel} shows the results of this comparison, where using multi-channel correlation map yields significantly higher results compared using a single-channel correlation map yielded by either Single$_{\textrm{concat}}$ or Single$_{\textrm{mean}}$.

\section{Additional qualitative results}
\begin{figure}
    \centering    
    \begin{center}
        \includegraphics[width=0.75\linewidth]{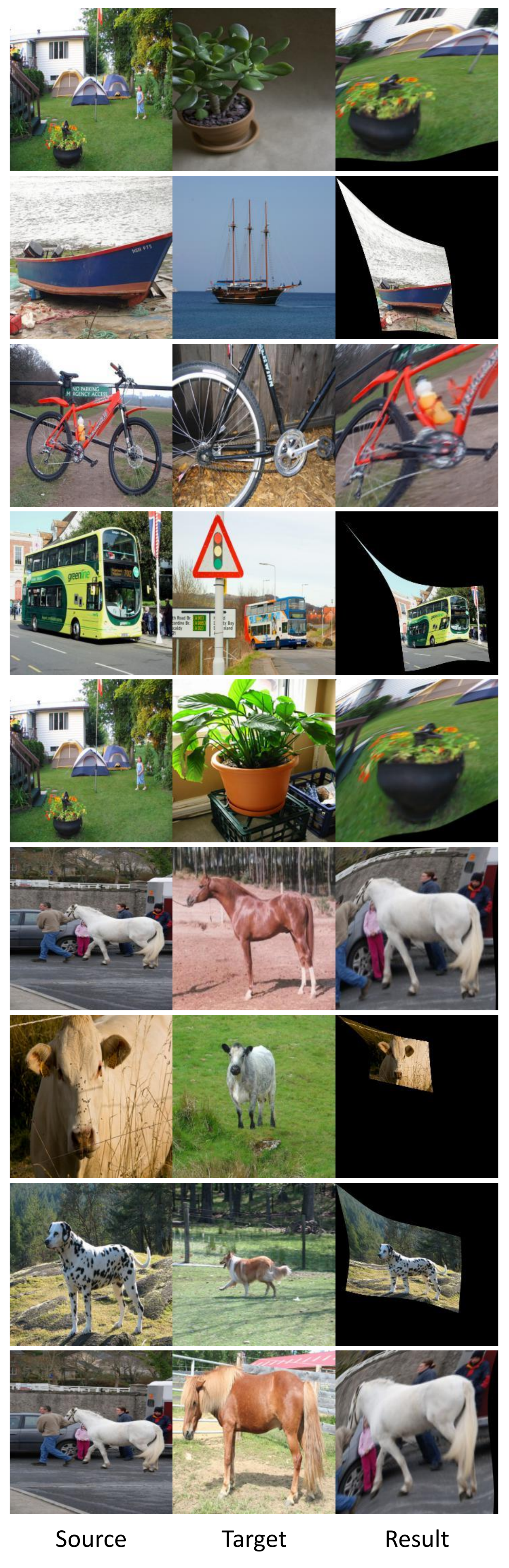}
    \end{center}
    \vspace{-5.0mm}
      \caption{Example visualization results with large scale changes from SPair-71k~\cite{min2019spair}.
}

    \vspace{-5.0mm}
\label{fig:scale_changes}
\end{figure}

\begin{figure}
    \centering    
    \begin{center}
        \includegraphics[width=0.75\linewidth]{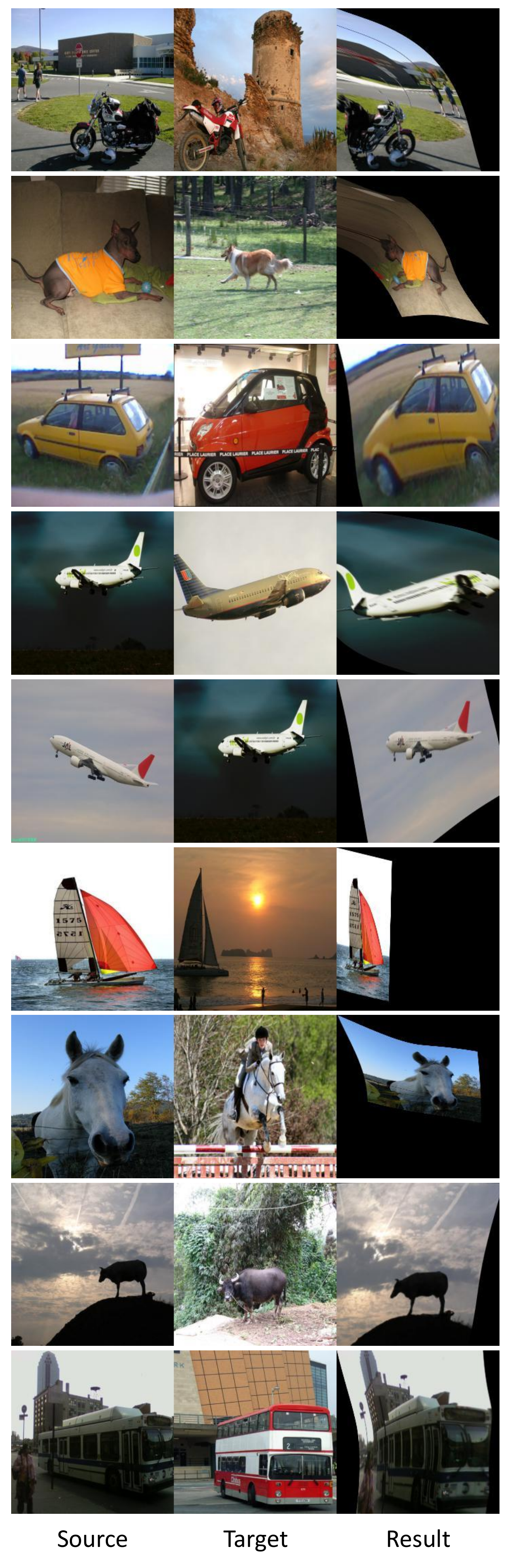}
    \end{center}
    \vspace{-5.0mm}
      \caption{Example visualization results with large viewpoint and illumination changes from SPair-71k~\cite{min2019spair}.
}
    \vspace{-5.0mm}
\label{fig:illu_vp_changes}
\end{figure}

\begin{figure*}
    \begin{center}
        \includegraphics[width=0.95\linewidth]{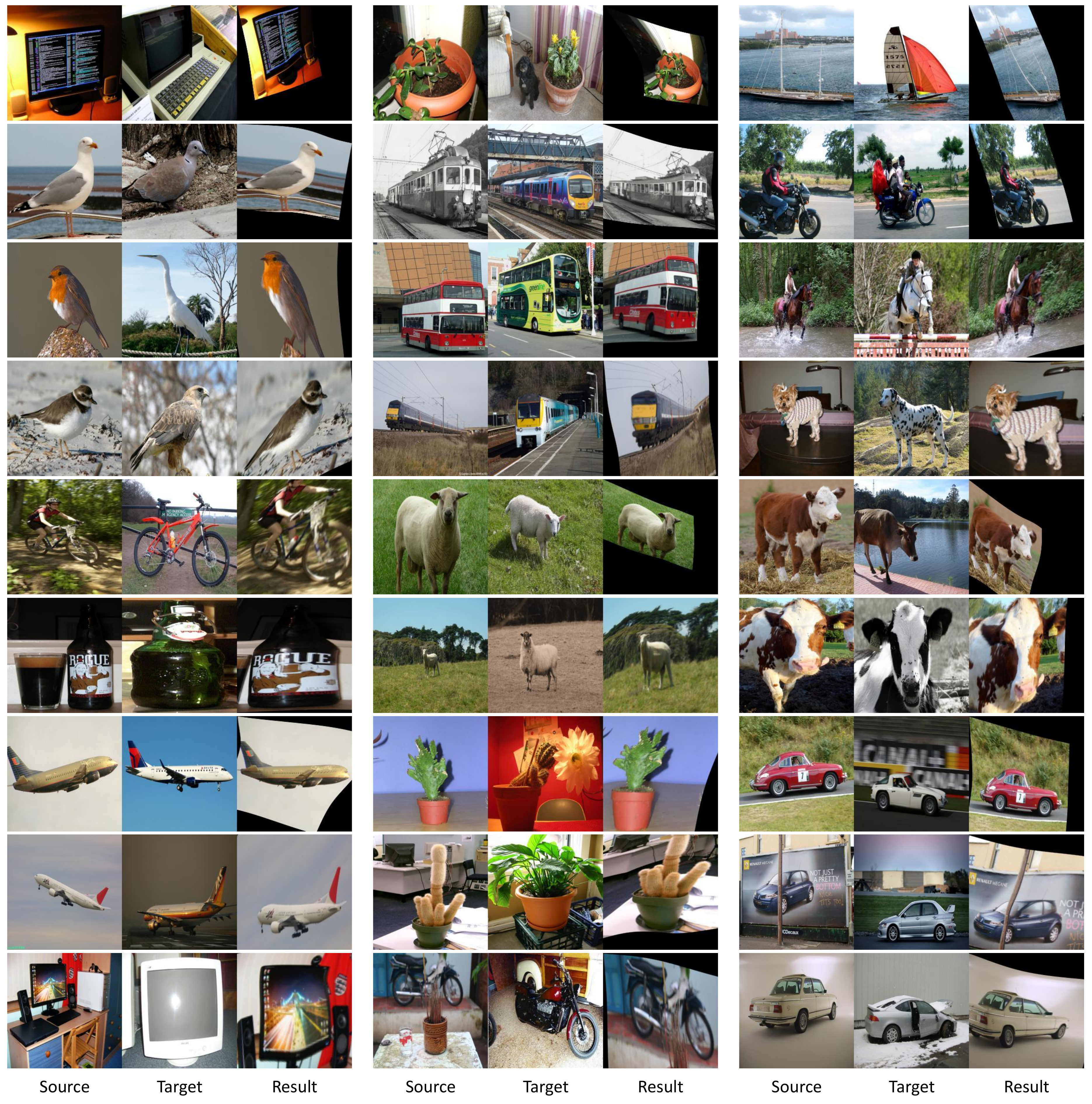}
    \end{center}
    \vspace{-3.0mm}
      \caption{Example visualization results from SPair-71k~\cite{min2019spair}.
    }
    \label{fig:sample_results}
    \vspace{-4.0mm}
\end{figure*}

In Fig.~\ref{fig:vis_comp}, we qualitatively compare TransforMatcher and CATs~\cite{cho2021semantic}, where TransforMatcher is seen to establish more accurate correspondences.
We also show additional example visualization results in Figures~\ref{fig:scale_changes}-\ref{fig:sample_results},
where the source image is TPS-transformed~\cite{donato2002approximate} to the target image using predicted correspondences, aligning common instances in each image pair.
As seen in Figures~\ref{fig:scale_changes} and \ref{fig:illu_vp_changes}, the proposed method, TransforMatcher, effectively aligns foreground instances in presence of large scale, viewpoint, and illumination differences.

\noindent
\section{Details on nonlocality analysis of match-to-match attention}
In this section, we provide implementation details regarding the analysis on nonlocality of match-to-match attention which is presented in the final part of \blue{section 5.2} of the main paper.
Recall that we define the measure of nonlocality of an MHSA at layer $l$ as the average of interactions between attention scores and relative offsets:
\begin{small}
\begin{align}
    \label{eqn:nonlocality_supp}
    \Phi^{l} = \frac{1}{Z} \sum_{h \in [N_h]} \sum_{(\mathbf{q}, \mathbf{k}) \in \mathcal{X} \times \mathcal{X}} \mathbf{A}_{\mathbf{q}, \mathbf{k}}^{(h)} \|\mathbf{q}-\mathbf{k}\|^{2}, 
\end{align}
\end{small}
where $Z$ is normalization constant and $\mathcal{X}$ is a set of spatial positions in $\mathbf{C}$.
As we found that the {\em global} query-key interaction in \blue{Eq.(5)} is inadequate to effectively quantify this metric, we build {\em pair-wise} query-key interaction: $\mathbf{A}_{\mathbf{q}, \mathbf{k}}^{(h)} = \sigma (\hat{\mathbf{Q}}^{(h)} \mathbf{K}^{(h)\top}) \in \mathbb{R}^{T \times T}$ where $\hat{\mathbf{Q}}^{(h)}_{i} \coloneqq \mathbf{Q}^{(h)}_{i} \sigma (\tau \mathbf{w}_{q} \mathbf{Q}^{(h)\top})$, $\mathbf{q}, \mathbf{k} \in \mathbb{R}^{4}$, and $T=HWHW$. The further the query attends ($\|\mathbf{q}-\mathbf{k}\|$), the larger the nonlocality ($\Phi^{l}$).

To measure the nonlocality of a convolutional layer, following the work of Cordonnier \etal ~\cite{cordonnier2020iclr}, we represent a $d$-dim conv layer with kernel size $K$ as an MHSA with $K^{d}$ heads with following constraint: $\sigma(\mathbf{A}^{(h)}_{\mathbf{q},:})_{\mathbf{k}}$ equals to $1$ if $\mathbf{q} - \mathbf{k} = \Delta_{K}$, and $0$ otherwise where $\Delta_{K}$ is a set of local offsets. For example, $\Delta_{K} \coloneqq [-1, 0, 1] \times [-1, 0, 1]$ if $d=2$ and $K=3$. We used $d \in \{4, 6\}$ and $K \in \{3, 5, 7, 9, 11\}$ in our experiments to visualize \blue{Figure 6}.

In plotting \blue{Figure 7} of the main paper, we utilize the difficulty levels of image pairs in the SPair-71k dataset. 
Each pair in SPair-71k has annotations describing the types (viewpoint \& scale variations, truncation, and occlusion) and levels (easy, medium, and hard) of difficulty. For truncation and occlusion, a pair is easy if no instances are truncated/occluded, medium if only one instance is, and hard if both are.

%% file: tables/supp_classwise.tex
\begin{table*}[!htb]
\centering
\scalebox{0.7}{
\begin{tabular}{lccccccccccccccccccc}
                \toprule
                Methods & aero & bike & bird & boat & bottle & bus & car & cat & chair & cow & dog & horse & mbike & person & plant & sheep & train & tv & all \\

                 \midrule
                 
                NC-Net~\cite{rocco2018neighbourhood} & 23.4 & 16.7 & 40.2 & 14.3 & 36.4 & 27.7 & 26.0 & 32.7 & 12.7 & 27.4 & 22.8 & 13.7 & 20.9 & 21.0 & 17.5 & 10.2 & 30.8 & 34.1 & 20.6 \\

                HPF~\cite{min2019hyperpixel} & 25.2 & 18.9 & 52.1 & 15.7 & 38.0 & 22.8 & 19.1 & 52.9 & 17.9 & 33.0 & 32.8 & 20.6 & 24.4 & 27.9 & 21.1 & 14.9 & 31.5 & 35.6&  28.2\\
                 
                SCOT~\cite{liu2020semantic} & 34.9 & 20.7 & 63.8 & 21.1 & 43.5 & 27.3 & 21.3 & 63.1 & 20.0 & 42.9 & 42.5 & 31.1 & 29.8 & 35.0 & 27.7 & 24.4 & 48.4 & 40.8 & 35.6\\
                 
                DHPF~\cite{min2020dhpf} & 38.4 & 23.8 & 68.3 & 18.9 & 42.6 & 27.9 & 20.1 & 61.6 & 22.0 & 46.9 & 46.1 & 33.5&  27.6&  40.1&  27.6&  28.1&  49.5&  46.5&  37.3\\
                 
                CHMNet~\cite{min2021chm}   & 49.6 & 29.3 & 68.7 & 29.7 & 45.3 & 48.4 & 39.5 & 64.9 & 20.3 & 60.5 & 56.1 & 46.0 &  33.8&  44.2&  38.9&  31.3&  72.2&  55.6&  46.4 \\
                 
                PMNC~\cite{lee2020pmnc} & 54.1 & \underline{35.9} & \textbf{74.9} & 36.5 & 42.1 & 48.8 & 40.0 & \textbf{72.6} & 21.1 & \textbf{67.6} & \textbf{58.1} & 50.5 & 40.1 & \textbf{54.1} & \textbf{43.3} & \textbf{35.7} & \underline{74.5}  & 59.9 & \underline{50.4}\\
                 
                MMNet~\cite{zhao2021multi} & 43.5 & 27.0 & 62.4 & 27.3 & 40.1 & 50.1 & 37.5 & 60.0 & 21.0 & 56.3 & 50.3 & 41.3 & 30.9 & 19.2 & 30.1 & 33.2 & 64.2 & 43.6 & 40.9\\
                 
                CATs~\cite{cho2021semantic}   & 46.5 & 26.9 & 69.1 & 24.3 & 44.3 & 38.5 & 30.2 & 65.7 & 15.9 & 53.7 & 52.2 & 46.7 & 32.7 & 35.2 & 32.2 & 31.2 & 68.0 & 49.1 & 42.4\\
                 
                CATs$\dagger$~\cite{cho2021semantic}   & 52.0 & 34.7 & 72.2 & 34.3 & \underline{49.9} & \underline{57.5} & 43.6 & 66.5 & 24.4 & 63.2 & 56.5 & \underline{52.0} & \underline{42.6} & 41.7 & 43.0 & 33.6 & 72.6 & 58.0 & 49.9\\
                 
                 \midrule
                 \midrule
                 TransforMatcher & \underline{54.5} & 33.9 & 72.2 & \underline{38.5} & 47.7 & 55.3 & \underline{45.6} & 65.7 & \underline{25.2} & 62.6 & \underline{58.0} & 47.0 & 40.7 & \underline{44.2} & \underline{43.1} & \underline{35.3} & 71.9 & \underline{61.6} & 50.2 \\
                 TransforMatcher$\dagger$  & \textbf{59.2} & \textbf{39.3} & \underline{73.0} & \textbf{41.2} & \textbf{52.5} & \textbf{66.3} & \textbf{55.4} & \underline{67.1} & \textbf{26.1} & \underline{67.1} & 56.6 & \textbf{53.2} & \textbf{45.0} & 39.9 & 42.1 & \underline{35.3} & \textbf{75.2} & \textbf{68.6} & \textbf{53.7} \\
                 \bottomrule
        \end{tabular}
        }
        \vspace{-1.0mm}
        \caption{\textbf{Classwise PCK on SPair-71k.} Higher PCK is better. All the results reported in the table uses pretrained ResNet-101 model as the feature extractor. $\dagger$ indicates the use of data augmentation during training.
        Numbers in bold indicate the best performance, followed by the underlined numbers.
        It can be seen that while TransforMatcher achieves the highest PCK overall, the usage of augmentation results in a decrease in PCK in certain categories.
        }
        \label{tbl:supp_classwise}
        \vspace{-3.0mm}
\end{table*}

%% file: tables/supp_multichannel.tex
\begin{table}[!htb]
\centering
\scalebox{1.0}{
\begin{tabular}{lcc}
                \toprule
                \multirow{3}{*}{Channel} & \multicolumn{2}{c}{SPair-71k} \\
                                         & \multicolumn{2}{c}{@$\alpha_{\text{bbox}}$} \\
                                         &  0.05 (F) & 0.1 (F) \\

                 \midrule
                 Single$_{\textrm{concat}}$ & 20.9  & 41.7 \\
                 Single$_{\textrm{mean}}$ & 24.1 & 45.1 \\
                 Multi (ours) & \textbf{32.4} & \textbf{53.7} \\
                 \bottomrule
        \end{tabular}
        }
        \vspace{-1.0mm}
        \caption{\textbf{Ablation on correlation map channel dimension.} Single$_\textrm{concat}$ and  Single$_{\textrm{mean}}$ denote single-channel correlation maps obtained by (1) concatenating the multi-layer features along the channel dimension prior to correlation computation, or (2) taking the mean of the multi-channel correlation map, respectively. 
        Using multi-channel correlation map yields the highest results.
        }
        \label{tbl:supp_channel}
        \vspace{-1.0mm}
\end{table}